\documentclass{article}
\usepackage{spconf,amsmath,graphicx}
\usepackage{xcolor}
\usepackage[hyphens]{url}
\usepackage{booktabs}
\usepackage{multirow}
\usepackage{makecell}
\usepackage{tabularx}
\usepackage{adjustbox}
\usepackage{cite}
\usepackage{subcaption}

\title{StyleBench: Evaluating Speech Language Models on Conversational Speaking Style Control} 
\name{
Haishu Zhao\textsuperscript{\rm 1},
Aokai Hao\textsuperscript{\rm 1},
Yuan Ge\textsuperscript{\rm 1},
Zhenqiang Hong\textsuperscript{\rm 1},
Tong Xiao\textsuperscript{\rm 1,2\textdagger}\thanks{\textdagger~Corresponding author},
Jingbo Zhu\textsuperscript{\rm 1,2}
}
\address{
\textsuperscript{\rm 1}NLP Lab, School of Computer Science and Engineering,
    Northeastern University, Shenyang, China\\
    \textsuperscript{\rm 2}NiuTrans Research, Shenyang, China\\
}
\begin{document}
%
\maketitle
\begin{abstract}
Speech language models (SLMs) have significantly extended the interactive capability of text-based Large Language Models (LLMs) by incorporating paralinguistic information. For more realistic interactive experience with customized styles, current SLMs have managed to interpret and control speaking style intensity from user prompts during the dialogue process. However, there remains a lack of systematic benchmarks that quantifies and evaluates the style intensity control ability in conversations. In this paper, we propose \textbf{StyleBench}\footnote{\textbf{StyleBench} datasets will be released after the paper is accepted.}, a multi-turn dialogue benchmark for comprehensively evaluating the style intensity control ability across four dimensions: emotion, speed, volume, and pitch. Our results reveal the performance gaps between leading SLMs and omni language models (OLMs), suggesting the underlying reasons and promising approaches for future exploration.
\end{abstract}
\section{Introduction}
\label{sec:intro}
Speech language models (SLMs) have recently demonstrated remarkable performance in reasoning and interactive scenarios \cite{arora2025landscapespokenlanguagemodels,peng2025surveyspeechlargelanguage,cui2025recentadvancesspeechlanguage}. Compared with text-based LLMs \cite{liang2024controllabletextgenerationlarge} that focus on purely semantic information, SLMs provide extra paralinguistic information and style characteristics, which significantly enhance the the interactive experience. Building on the paradigm established by GPT-4o \cite{openai2024gpt4ocard}, a growing number of SLMs \cite{chen2025emovaempoweringlanguagemodels,défossez2024moshispeechtextfoundationmodel,chen2025minmomultimodallargelanguage} have demonstrated the ability to synthesize speech with variable style features and intensities. Some of these models, which are pre-trained on relevant style enhancement datasets\cite{zeng2024glm4voiceintelligenthumanlikeendtoend,kimiteam2025kimiaudiotechnicalreport}, can even dynamically control the style and intensity in multi-turn interactions. The advancements in stylistic speech generation have significantly enhanced the experience of human-computer interaction. 

\begin{figure}[t]
\centering
\includegraphics[width=\columnwidth]{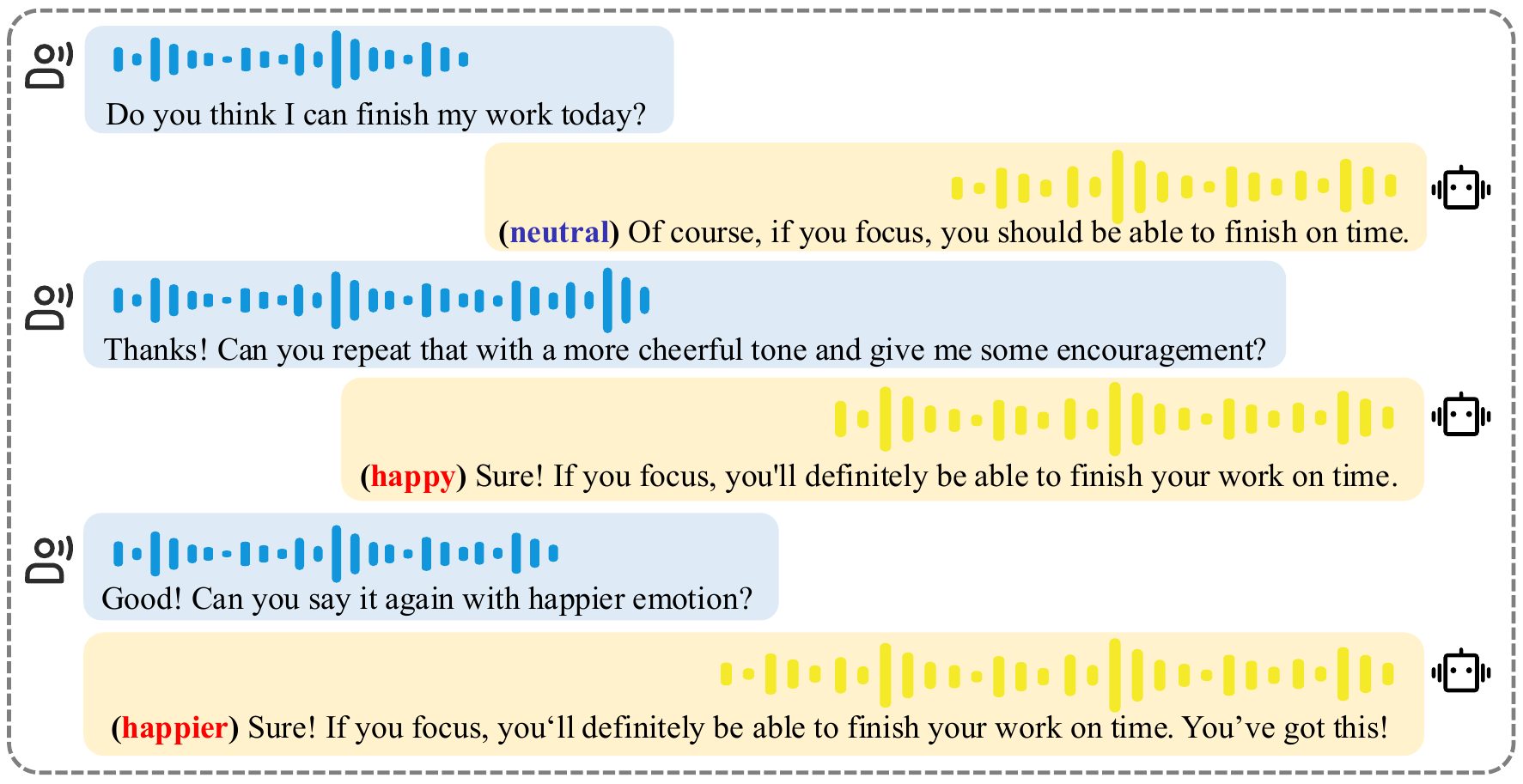} 
\caption{Conversational Speaking Style Controlling Dialogue}
\label{fig1}
\end{figure} 

Despite the advancements, the lack of a systematic benchmark to evaluate the speaking style intensity control ability remains a challenge. While we can sense that different SLMs have varying abilities in interpreting and expressing style features, it is still a non-trivial task assessing whether models faithfully follow the stylistic prompts (Fig.1). Previous works on benchmarking SLMs mostly focused on single-turn conversations and conventional tasks, such as ASR and spoken question answering tasks \cite{gao2025benchmarkingopenendedaudiodialogue,yang2024airbenchbenchmarkinglargeaudiolanguage,riviere2021asr4realextendedbenchmarkspeech}. More recently, AudioBench \cite{wang2025audiobenchuniversalbenchmarkaudio} and SpeechFeedback \cite{ge2025sagelm} extends previous work by adding an emotional evaluation benchmark, but it is still limited to distinguishing emotional categories, lacking multi-dimensional style evaluation and intensity variation quantification in multi-turn dialogues. 
Since we often need models to adjust the style and intensity of the response for a desired result during daily interaction, there is an urgent need of systematic benchmarks for evaluating style intensity variation in dialogue scenarios. 


\begin{figure*}[ht]
    \centering
    \begin{subfigure}[t]{0.3\linewidth} 
        \centering
        \includegraphics[height=4.2cm]{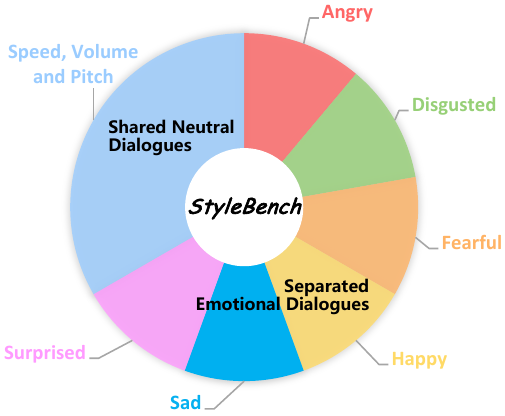} 
        \caption{\textbf{StyleBench} Data Composition}
    \end{subfigure}
    \hspace{0.01\linewidth} %
    \begin{subfigure}[t]{0.65\linewidth}
        \centering
        \includegraphics[height=4.2cm]{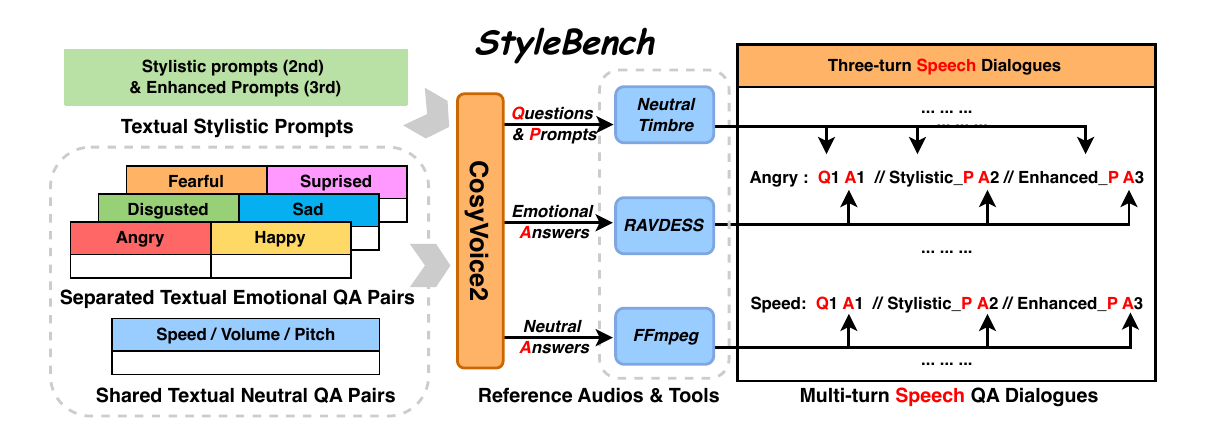} 
        \caption{\textbf{StyleBench} Data Synthesis Process}
    \end{subfigure}
    \caption{An overview of data composition and synthesis process: textual contents are generated with different situational features (emotional or neutral), followed by neutral prompts and stylistic responses synthesized using \textbf{\textit{CosyVoice2}}. Emotional responses were synthesized using the \textbf{\textit{RAVDESS}} as the reference audio, while the others were adjusted using \textbf{\textit{FFmpeg}}.}
    \label{fig:two_images}
\end{figure*}

In this paper, we present \textbf{StyleBench}, a systematic benchmark for evaluating  conversational speaking style control across four dimensions: Emotion, Speed, Volume and Pitch. Our main contributions are thus: (1) providing \textbf{StyleBench}, a comprehensive benchmark with 14.4K multi-turn QA dialogues for systematic evaluation on style control and intensity variation, (2) developing dimension-specific evaluation toolkits that focus on quantifying the control ability and style variation degree across dialogue turns, and (3) revealing substantial performance gaps between leading SLMs and OLMs through the evaluation of 10 end-to-end open-source SLMs. By revealing the differences in training datasets and speech tokenizers, \textbf{StyleBench} provides some insights for advanced stylistic speech conversation.


\section{Benchmark Dataset Construction}
\label{sec:Task Setup and Adaption}
Evaluating the conversational speaking style control ability of SLMs necessitates coherent multi-turn dialogue modeling alongside accurate stylistic speech generation. We therefore construct a multi-turn QA dataset designed for controlling style and intensity. In this section, we outline \textbf{StyleBench}, a bilingual three-turn dialogue dataset that enables the evaluation of SLMs on: (1) understanding natural language instructions with gradational style variations, and (2) preserving hierarchical intensity control across dialogue turns.

\subsection{Data Samples}
In contrast to previous multi-turn datasets \cite{koudounas2025deepdialoguemultiturnemotionallyrichspoken, yang2022opensourcemagicdataramcrich}, each data sample in \textbf{StyleBench} is organized as a three-turn dialogue with progressively varying style intensity. The first turn serves as a neutral baseline, where the model generates a default response. In the second and third turns, the same utterance is re-synthesized under increasingly strong style-related instructions, forming either an \textbf{\textit{intensifying}} or an \textbf{\textit{attenuating}} trajectory. Furthermore, user prompts are phrased in conversational forms rather than templated commands or categorical labels. Distinct prompts are designed for each style dimension, ensuring coherent dialogue scenarios. 

Each sample controls only a single dimension at a time while keeping semantic content fixed. All utterances are synthesized independently with CosyVoice2 \cite{du2024cosyvoice2scalablestreaming}, ensuring that the intensity variation arises solely from the prompts. This design isolates the effect of prompt-based supervision and provides a clear measure of a model’s controllability across multiple style dimensions. Table 1 and Fig.2 (a) illustrate the dataset size and data composition. 

\begin{table}[t]
\centering
\begin{adjustbox}{max width=\linewidth}
\begin{tabular}{l|c|c|c}
\toprule
\multicolumn{1}{c}{\textbf{Style}} & \multicolumn{1}{c}{\textbf{Speakers}} & \multicolumn{1}{c}{\textbf{Utterances}} & \multicolumn{1}{c}{\textbf{Duration(h)}} \\
\midrule 
Emotion & 8 (4en/4zh) & 57600 & 84.88 \\
Speed & 8 (4en/4zh) & 9600 & 11.76 \\
Volume & 8 (4en/4zh) & 9600 & 9.54 \\
Pitch & 8 (4en/4zh) & 9600 & 9.35 \\
\bottomrule
\end{tabular}
\end{adjustbox}
\caption{Overview of \textbf{StyleBench} dataset.}
\end{table}

\subsection{Textual Content} 
For the emotion subsets, each QA pair is designed to convey the contextual signals of the target emotion while deliberately avoiding explicit affective terms, thereby preserving naturalness and preventing label leakage. For Speed, Volume and Pitch, we adopt a shared pool of semantically neutral dialogues. These dialogues are reused across different prosodic dimensions to ensure that evaluation focuses exclusively on acoustic variations. In the second and third turns of each dialogue, we introduce targeted prompting instructions that either amplify or attenuate the style intensity. Crucially, the textual answers remain identical across turns, ensuring any variation is attributable solely to paralinguistic modulation.

\subsection{Speech Synthesis}
While all speech samples in the dataset are synthesized using CosyVoice2 (Fig.2) \cite{du2024cosyvoice2scalablestreaming}, intensity variation is introduced through differences in \textbf{reference audio} and \textbf{post-processing procedures}. 
For the answers of emotion subset, we use reference audio from \textbf{\textit{RAVDESS}}\footnote{\url{ https://github.com/tuncayka/speech_emotion dataset}}, which contains high-quality English speech across various emotions and intensities. Accordingly, each sample is rendered into a three-turn answers with different emotional intensities. For the other dimensions, we first synthesize base utterances in a neutral style, which are then processed using \textbf{\textit{FFmpeg}}\footnote{\url{ https://github.com/FFmpeg/FFmpeg}} to produce controlled variations with different intensities. By varying the order of QA pairs, we organize three-turn sequences that represent either increasing or decreasing intensity. Meanwhile, we randomly assign eight responder timbres across samples in each subset to increase the timbre diversity. 

\section{Evaluations \& Discussions}
\label{sec:Evaluation}
Building on the \textbf{StyleBench} dataset, we conduct a comprehensive evaluation of mainstream SLMs, providing a systematic characterization of their speaking style control performance. Based on the evaluation results, we further analyze the underlying causes of the obvious performance gap among various SLMs through fine-grained experiments.

\subsection{Style Quantification Metrics}
Emotional classification tasks are usually automatically completed by classification models like Emotion2Vec \cite{ma2023emotion2vecselfsupervisedpretrainingspeech}. However, their reliability drops on evaluating the style intensities within synthesized data. As an auxiliary measure, softmax confidence score also quickly saturates (often $>$ 0.95) and fails to capture fine-grained changes across turns. 
Hence, we incorporate human evaluation to judge whether emotional intensity varies as expected.

For the others, speed is quantified as syllables per minute (\textbf{SPM}) based on Whisper-large-v3\footnote{\url{https://huggingface.co/openai/whisper-large-v3}} transcriptions. Volume is measured by the root mean square (\textbf{RMS}) energy of the waveform, reflecting perceived loudness. Pitch follows the definition in FastSpeech2 \cite{ren2022fastspeech2fasthighquality}, computed as the average fundamental frequency (\textbf{F0}) over voiced frames. The framework integrates automatic metrics with human evaluation to capture the style species and intensity variations across dimensions.

\subsection{Experimental Setup}
We introduce a three-stage evaluation strategy for mainstream SLMs. The first stage examines the instruction-following ability in \textbf{single-turn dialogues}. The second stage extends this to \textbf{multi-turn dialogues}, evaluating whether the model can maintain consistent instruction-following across turns. The third stage further evaluates the \textbf{speaking style control ability}. For all experiments, 10\% of the dataset is reserved for the following evaluation.

Semantic relevance in the first two stages is measured by Qwen3-4B-Instruct \cite{yang2025qwen3technicalreport}, which computes the \textbf{relevance degree} between questions and answers. For conversational style control, we adopt a two-step evaluation. First, we measure the probability of producing a valid response, defined as the \textbf{V}alid \textbf{S}ample \textbf{P}ercentage (\textbf{VSP}). A sample is considered valid if the model produces a \textbf{distinct} and \textbf{intended} stylistic output in line with the prompt. Second, we compute the \textbf{S}tyle \textbf{V}ariation \textbf{D}egree (\textbf{SVD}) for all valid samples in quantifiable dimensions (Speed, Volume and Pitch). Let $S_{T1}$, $S_{T2}$, and $S_{T3}$ denote the style scores at each turn. \textbf{SVD} ($\Delta_1$$\mid$$\Delta_2$) is defined as:
\[
\Delta_1 = \left| \frac{S_{T2} - S_{T1}}{S_{T1}} \right| \times 100\%, \quad
\Delta_2 = \left| \frac{S_{T3} - S_{T2}}{S_{T2}} \right| \times 100\%
\]
Here, the absolute percentage difference captures the magnitude of style adjustment, independent of direction. By jointly analyzing \textbf{VSP} and \textbf{SVD}, we provide a quantitative characterization of speaking style controlling capability.

\begin{table}[t]
\centering
\begin{adjustbox}{max width=\linewidth}
\begin{tabular}{l|c|c|c}
\toprule
\multicolumn{1}{c}{\textbf{Model}} & \multicolumn{1}{c}{\textbf{Size}} & \multicolumn{1}{c}{\textbf{SRD}(\%)$\uparrow$} & \multicolumn{1}{c}{\textbf{MRD}(\%)$\uparrow$} \\
\midrule 
Mini-omni \cite{xie2024miniomnilanguagemodelshear} & 0.5B & 50.14 & \textcolor{red}{---}  \\
Mini-omni2 \cite{xie2024miniomni2opensourcegpt4ovision} & 0.5B & 62.78 & \textcolor{red}{---} \\
Slam-omni \cite{chen2024slamomnitimbrecontrollablevoiceinteraction} & 0.5B & 66.18 & \textcolor{red}{28.19} \\
Freeze-omni \cite{wang2024freezeomnismartlowlatency} & 7B & 91.94 & \textcolor{red}{26.46} \\
MiniCPM-o 2.6 \cite{hu2024minicpmunveilingpotentialsmall} & 7B & 95.07 & \textcolor{red}{26.18} \\
Qwen2.5-omni \cite{xu2025qwen25omnitechnicalreport} & 7B & \textbf{97.36} & 64.51 \\
Baichuan-omni-1.5 \cite{li2025baichuanomni15technicalreport} & 7B & 89.03 & 55.76 \\
LLaMA-omni2 \cite{fang2025llamaomni2llmbasedrealtimespoken} & 8B & 91.94 & 44.03 \\
GLM-4-Voice \cite{zeng2024glm4voiceintelligenthumanlikeendtoend} & 9B & 91.53 & \textbf{69.31} \\
Kimi-Audio \cite{kimiteam2025kimiaudiotechnicalreport} & 7B & 90.62 & 67.43 \\
\bottomrule
\end{tabular}
\end{adjustbox}
\caption{Average Semantic \textbf{R}elevance \textbf{D}egree in \textbf{S}ingle-turn (SRD) \& \textbf{M}ulti-turn (MRD) dialogues.}
\end{table}

\begin{table*}[t]
\centering
\begin{adjustbox}{max width=\linewidth}
\begin{tabular}{l|c|c|c|c|c|c}
\toprule
\multicolumn{1}{c}{\multirow{2.5}{*}{\textbf{Model}}} & 
\multicolumn{6}{c}{\textbf{VSP}(\%)$\uparrow$} \\ 
\cmidrule(lr){2-7}
 \multicolumn{1}{c}{} & \multicolumn{1}{c}{\textbf{Angry}} & \multicolumn{1}{c}{\textbf{Disgusted}} & \multicolumn{1}{c}{\textbf{Fearful}} & \multicolumn{1}{c}{\textbf{Happy}} & \multicolumn{1}{c}{\textbf{Sad}} & \multicolumn{1}{c}{\textbf{Surprised}} \\
\midrule
LLaMA-omni2 & ---~~~ $\mid$ ~~~--- & ---~~~ $\mid$ ~~~--- & ---~~~ $\mid$ ~~~--- & ---~~~ $\mid$ ~~~--- & ---~~~ $\mid$ ~~~--- & ---~~~ $\mid$ ~~~--- \\
Baichuan-omni-1.5 & ---~~~ $\mid$ ~~~--- & 12.50 $\mid$ 11.25 & ---~~~ $\mid$ ~~~--- & 25.00 $\mid$ 25.00 & ---~~~ $\mid$ ~~~--- & 19.38 $\mid$ ~~8.13 \\
Qwen2.5-omni & 23.13 $\mid$ 13.75 & 36.88 $\mid$ 20.63 & 11.25 $\mid$ ~~6.25 & 40.00 $\mid$ 30.00 & 24.38 $\mid$ 18.13 & 29.38 $\mid$ 23.13 \\
GLM-4-Voice & 50.63 $\mid$ \textbf{36.88} & 38.75 $\mid$ \textbf{43.13} & 41.25 $\mid$ \textbf{26.25} & 44.38 $\mid$ \textbf{33.13} & 57.50 $\mid$ \textbf{51.25} & 36.88 $\mid$ \textbf{45.63} \\
Kimi-Audio & \textbf{68.75} $\mid$ 15.63 & \textbf{76.25} $\mid$ 37.50 & \textbf{55.00} $\mid$ 10.00 & \textbf{47.50} $\mid$ 21.25 & \textbf{73.13} $\mid$ 34.38 & \textbf{53.75} $\mid$ 26.88 \\
\bottomrule
\end{tabular}
\end{adjustbox}
\caption{\textbf{V}alid \textbf{S}ample \textbf{P}ercentage (\textbf{VSP}) is used to measure the proportion of test cases where the model exhibited a \textbf{distinct} and \textbf{intended} response (Turn$\_2$ $\mid$ Turn$\_3$). "---" indicates that the model is completely unresponsive to emotional adjustment.}
\end{table*}

\begin{table*}[t]
\centering
\begin{adjustbox}{max width=\linewidth}
\begin{tabular}{l|ccc|ccc}
\toprule
\multicolumn{1}{c}{\multirow{2.5}{*}{\textbf{Model}}} &
\multicolumn{3}{c}{\textbf{VSP}(\%)$\uparrow$} & 
\multicolumn{3}{c}{\textbf{SVD}(\%)$\uparrow$} \\
\cmidrule(lr){2-4} \cmidrule(lr){5-7}
\multicolumn{1}{c}{} & \textbf{Speed} & \textbf{Volume} & \multicolumn{1}{c}{\textbf{Pitch}}
& \multicolumn{1}{c}{\textbf{Speed}} & \textbf{Volume} & \textbf{Pitch} \\
\midrule
LLaMA-omni2 & 50.00 $\mid$ 49.38 & 50.00 $\mid$ 41.25 & 55.00 $\mid$ 46.88 & \ \ 9.50 $\mid$ 16.95 & 17.56 $\mid$ 17.17 & 4.54 $\mid$ 5.11 \\
Baichuan-omni-1.5 & 48.75 $\mid$ 46.25 & 48.75 $\mid$ \textbf{58.13} & 46.25 $\mid$ 45.00 & 13.67 $\mid$ 12.99 & 13.09 $\mid$ 11.02 & 5.63 $\mid$ 5.99 \\
Qwen2.5-omni & 52.50 $\mid$ 50.62 & 46.25 $\mid$ 50.62 & 52.50 $\mid$ \textbf{50.62} & ~~8.35 $\mid$ ~~6.78 & ~~7.50 $\mid$ ~~7.79 & 5.50 $\mid$ 5.11\\
GLM-4-Voice & 77.50 $\mid$ 71.88 & \textbf{61.25} $\mid$ 49.38 & \textbf{68.12} $\mid$ 50.00 & 19.38 $\mid$ 14.75 & \textbf{31.96} $\mid$ \textbf{19.27} & 7.58 $\mid$ 4.07\\
Kimi-Audio & \textbf{81.88} $\mid$ \textbf{78.75} & 53.12 $\mid$ 50.00 & 61.88 $\mid$ 44.38 & \textbf{29.94} $\mid$ \textbf{22.26} & 17.94 $\mid$ 14.98 & \textbf{10.10} $\mid$ \textbf{8.43}\ \ \ \\
\bottomrule
\end{tabular}
\end{adjustbox}
\caption{\textbf{V}alid \textbf{S}ample \textbf{P}ercentage (\textbf{VSP}) is used to measure the proportion of test cases where the model exhibited a \textbf{distinct} and \textbf{intended} response (Turn$\_2$ $\mid$ Turn$\_3$). For all valid samples, \textbf{S}tyle \textbf{V}ariation \textbf{D}egree (\textbf{SVD}) represents the corresponding style intensity shifts in consecutive turns ($\Delta_1$ $\mid$ $\Delta_2$).}
\end{table*}

\subsection{Main Results}
We first evaluate these SLMs on semantic instruction-following using Single-turn Relevance Degree (\textbf{SRD}) and Multi-turn Relevance Degree (\textbf{MRD}). SRD measures adherence to instructions in isolated QA pairs, whereas MRD captures semantic coherence across dialogue turns. Since speaking style controlling relies on the ability to maintain consistent dialogue context, MRD serves as the primary prerequisite for subsequent evaluation. As shown in Table~2, while most large-scale models achieve high SRD, their MRD performance varies significantly. Only Qwen2.5-omni \cite{xu2025qwen25omnitechnicalreport}, GLM-4-Voice \cite{zeng2024glm4voiceintelligenthumanlikeendtoend}, and Kimi-Audio \cite{kimiteam2025kimiaudiotechnicalreport} exceed 60\%, indicating reliable multi-turn consistency.

Accordingly, we apply a threshold of MRD $>$ 40\% to filter eligible models for the evaluation on speaking style control, which ensures a minimal level of multi-turn coherence for more reliable results. On this basis, we manually evaluate the performance of the different SLMs on the emotional dimension. As reported in Table 3, Kimi-Audio has a leading edge in every emotion category. Notably, we also find that Kimi-Audio exhibits distinct emotional features during the initial adjustment, which consequently diminishes the effectiveness in the following turn (fall behind GLM-4-Voice in Turn 3). In contrast, since OLMs mostly focus on spoken QA and speech instruction following tasks, LLaMA-omni2 and Baichuan-omni-1.5 show almost no response to the user instructions for emotional adjustment.

As presented in Tables 4, Kimi-Audio and GLM-4-Voice consistently demonstrate stronger style control ability in Speed, Volume and Pitch. They achieve higher VSP and SVD, suggesting not only a higher probability of producing distinct stylistic responses but also distinct intensity expression to style prompts. In contrast, the others often fail to generate valid responses or exhibit unconspicuous style variation, reflecting limited control ability.

\subsection{Discussions}
Although these SLMs are relatively comparable in size, their performances have shown substantial divergence in each dimension. Hence, such variation suggests other critical factors inside the models. In this section, we analyze some important factors that impact the performance on speaking style control.

\subsubsection{\textbf{Impact of Data Training}}
We find that the training datasets of the underperforming omni-models are predominantly curated for conventional tasks such as ASR and Spoken question answering. In contrast, GLM-4-Voice \cite{zeng2024glm4voiceintelligenthumanlikeendtoend} combines unsupervised speech datasets during pre-training process, enabling the model to acquire stylistic patterns from natural dialogues. Furthermore, Kimi-Audio \cite{kimiteam2025kimiaudiotechnicalreport} introduces a dataset explicitly designed to enhance the speaking style control ability. These differences highlight the critical influence of training data on shaping such abilities.


\subsubsection{\textbf{Impact of Speech Tokenizers}}
In addition to training the inference ability, we also focus on how the model expresses the acoustic information. Hence, we manage to analyze the influence of speech \textbf{tokenizer} on the speech synthesis. While most of the SLMs typically employ a flow-matching based decoder \cite{du2024cosyvoicescalablemultilingualzeroshot} to convert speech tokens into waveforms, our analysis emphasizes the speech tokens themselves, which seems to be the cause of performance gaps. We find that the speech tokens inferred from the same textual content in different styles are different, suggesting that these tokens inherently preserve paralinguistic cues. GLM-4-Voice further validates this observation: compared with earlier designs such as SpeechTokenizer \cite{zhang2024speechtokenizerunifiedspeechtokenizer} and Whisper-large-v3, its independently trained tokenizer achieves superior retention of both semantic and acoustic information. These results underscore the crucial role of speech tokenizers in capturing and reproducing the style variations.

\section{Conclusion}
In this work, we present \textbf{StyleBench}, a systematic benchmark for evaluating the speaking style control ability. Leveraging a multi-turn dialogue dataset that spans four dimensions, \textbf{StyleBench} supports comprehensive evaluation in real-time conversations. Our results have revealed substantial gaps across models of advanced style control, suggesting prospective directions for further exploration.

\bibliographystyle{IEEEbib}
\bibliography{strings,refs}

\end{document}